\documentclass[letterpaper, 10 pt, conference]{ieeeconf}  

\usepackage{mathtools}
\usepackage{float}

\usepackage{amssymb}
\usepackage[ruled,vlined,linesnumbered]{algorithm2e}
\usepackage{amsmath}
\usepackage{amsfonts}
\usepackage{amssymb}
\usepackage[caption = false]{subfig}

\newcommand{\beqa}{\begin{eqnarray}}
\newcommand{\eeqa}{\end{eqnarray}}

\newcommand{\sgn}{\mathrm{sgn}}

\newcommand{\barr}{\begin{array}}
\newcommand{\earr}{\end{array}}

\newcommand\bbR{\mathbb{R}}

\newif\ifshowdetail\showdetailfalse

\newcommand{\bsbeq}{\begin{subequations}}
\newcommand{\esbeq}{\end{subequations}}

\IEEEoverridecommandlockouts                              

\title{\LARGE \bf
Proxy-based Super Twisting Control Algorithm for Aerial Manipulators}

\author{Zhengyu Hua, Bowen Xu, Li Xing, Fengyu Quan, Xiaogang Xiong, Haoyao Chen*
  \thanks{This work was supported in part by the National Natural Science Foundation of China under Grant U1713206 and Grant 61673131. (Corresponding author: Haoyao Chen, Xiaogang Xiong.)}
  \thanks{Z.Y. Hua, B.W. Xu, L. Xing, F.Y. Quan, X.G. Xiong, H.Y. Chen* are with the School of Mechanical Engineering and Automation, Harbin Institute of Technology Shenzhen, P.R. China, e-mail: hychen5@hit.edu.cn.}
}

\begin{document}

\maketitle
\thispagestyle{empty}
\pagestyle{empty}

\begin{abstract}

Aerial manipulators are composed of an aerial multi-rotor that is equipped with a 6-DOF servo robot arm. To achieve precise position and attitude control during the arm's motion, it is critical for the system to have high performance control capabilities. However, the coupling effect between the multi-rotor UAVs' movement poses a challenge to the entire system's control capability. We have proposed a new proxy-based super twisting control approach for quadrotor UAVs that mitigates the disturbance caused by moving manipulators. This approach helps improve the stability of the aerial manipulation system when carrying out hovering or trajectory tracking tasks. The controller's effectiveness has been validated through numerical simulation and further tested in the Gazebo simulation environment. \end{abstract}

\section{Introduction}

In recent years, the aerial manipulator system\cite{RuggieroSurvey2018} has garnered significant attention. Generally, this system comprises a robotic manipulator and a multi-rotor UAV, giving it the capability to actively interact with its surroundings by performing tasks like grasping and transportation. This is in contrast with traditional aerial automated systems that can only execute passive tasks like monitoring and surveillance. However, the strong coupling between the aerial vehicle and the robotic manipulator presents a significant challenge for precise control and manipulation. Specifically, the motion of the robot arm can lead to disturbances from the UAV perspective. Moreover, owing to the under-actuated nature of multi-rotor UAVs, it may be challenging for the UAV to correct the disturbance and achieve accurate tracking.
From a control perspective, the aerial manipulation system can be treated as two controlled objects with two separate controllers designed for aerial vehicles and manipulators. 
The dynamic effect caused by the motion of the manipulator on multi-rotor UAV is difficult to model, and thus it can be treated as external forces and torques \cite{ruggiero2014impedance, ruggiero2015passivity}. Therefore, most related works focused on UAV anti-disturbance control. A Variable Parameter Integral Back-stepping (VPIB)\cite{jimenez2013control} UAV control approach taking the motion of the arm into account, which outperforms the results of traditional cascaded PID controller. Based on a novel disturbances estimator, impedance\cite{ruggiero2014impedance} and passivity-based\cite{ruggiero2015passivity} control methods are implemented for UAV to compensate for the estimated disturbances. In addition, a combination of disturbance observer (DoB) and robust control approach\cite{kim2017robust} is proposed to deal with the external forces and model uncertainties. According to the dynamic model of UAV, a disturbance compensation robust $H_{\infty}$ controller combined with a disturbance estimator is designed to increase stability. At the same time, the aerial manipulator conducts hovering operation tasks.

Terminal sliding mode control (TSMC) is increasingly popular due to its ability to handle system uncertainties and external disturbances. While TSMC has not yet been implemented on aerial manipulators, several sliding mode control methods have been developed for multi-rotor UAVs to enhance robustness and stability, as shown in Xu's work \cite{xu2006sliding}. However, one issue with sliding mode control is the chattering phenomenon, which can have severe consequences when controlling multi-rotor UAVs. A conventional cascaded PID controller with an inner-loop for attitude control and an outer-loop for position control was proposed for UAVs in \cite{l2018introduction}; however, this approach cannot guarantee finite time convergence and produces a significant chattering effect. Recent works aim to reject external disturbances using sliding mode control and eliminate chattering effects. In \cite{mofid2020adaptive}, an adaptive PID-SMC technique is proposed for multi-rotor position tracking control. The method takes into account external disturbances and creates adaptive inconsistent inputs to suppress chattering on the sliding surface.

From the super-twisting control perspective, there are usually two types of strategies to attenuate the magnitudes of numerical chattering. The first strategy is in a continuous manner, including the adaptive gain schemes in \cite{Ding_2020_TCASI,Utkin_2013_ASTA} and the singularity-free methods \cite{Mobayen_2011_ACC,Mobayen_2017_Scientia}. The second strategy relies on discrete-time implementation, such as these schemes in \cite{Brogliato_2019_STA}. In \cite{Xiong_2019_TCASII}, implicit Euler discretizations of the twisting algorithm were proposed, and the numerical chattering was avoided. Implicit discrete-time homogeneous differentiators with recursive formulations of higher-order super-twisting algorithms were discussed in \cite{Brogliato_2020_differentiator} to achieve higher estimation accuracy with chattering suppressions. However, there are no implicit Euler realizations of TSMC with nonlinear sliding surfaces, although TSMC is widely employed \cite{Ding_2020_TCASI}. The difficulty lies in how to deal with the nonlinear sliding surfaces in TSMC. Inspired by the work in \cite{Brogliato_2020_differentiator}, to suppress the numerical chattering, the implicit Euler method is employed to realize the second-order TSMC in a discrete-time manner.

This paper aims to develop a novel approach to address the challenges posed by the model uncertainties and disturbances in aerial manipulators through the utilization of the super-twisting concept in multi-rotor attitude and position control. The remainder of the study is structured as follows. Section II outlines the general kinetic model and dynamic model of the quadrotor system. Section III proposes the Proxy-based Super-Twisting Algorithm (PSTA) and explains how it is implemented on the quadrotor. Section IV and V present the numerical and Gazebo simulation outcomes, respectively. Finally, Section VI offers conclusions and suggestions for future research.


\section{Modelling}
\begin{figure}[ht]
    \centering
    \includegraphics[width=6.4cm]{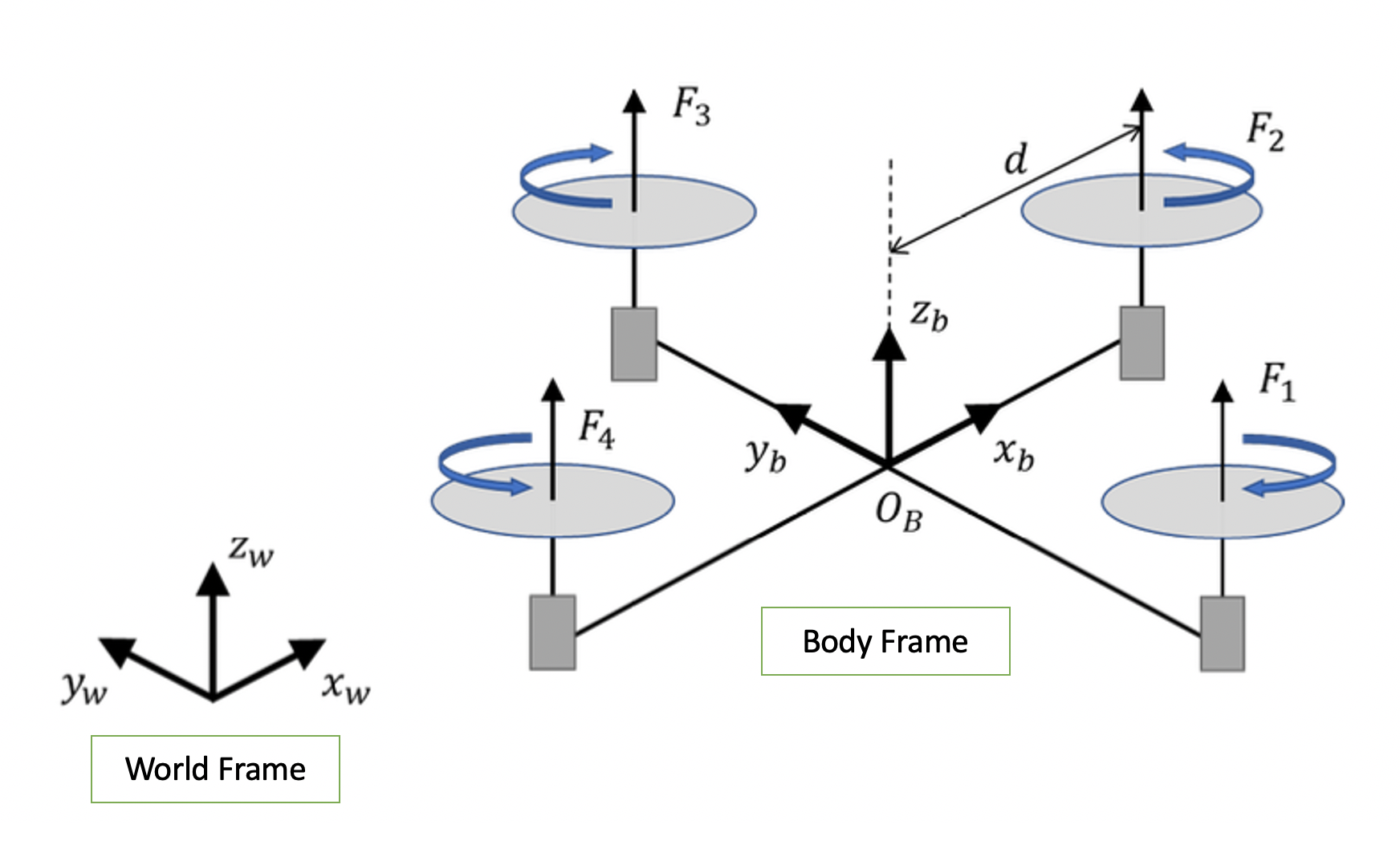}
    \caption{Modeling of MAV with movable perception system.}
    \label{model}
\end{figure}

This section presents the quadrotor model used in the controller formulation. A world-fixed inertial frame and body-fixed frame placed at the quadrotor center of mass are defined as shown in Fig. 1. The configuration of this vehicle is determined by the localization of the center of mass $p=[x,y,z]^T$ and the rotation matrix from the body to world frame ${R_{b}^{w}}$. The following equations can describe the quadrotor system dynamics:

\begin{equation}
\left\{\begin{array}{l}\dot{p}=v, \\ \dot{R_{b}^{w}}=R_{b}^{w}\hat{\omega},
 \\m \dot{v}=-m g e_{3}+f_{u} R_{b}^{w} e_{3}+f_{ext},
 \\ J \dot{\omega}=-\omega \times J \omega+M_{u}+M_{ext}.
\end{array}\right.
\end{equation}
where the unit vector $e_3=[0,0,1]^T$. $v$ is the velocity of the center of mass expressed in the world-fixed inertial frame; $g$ is the acceleration of gravity, $m$ is the total mass, and $J$ is the inertia matrix with respect to the body frame; $\omega$ is the angular velocity in the body-fixed frame; $f_{u}$ and $M_{u}$ are the control thrust and moment generated by the propellers, respectively; $f_{ext}$ and $M_{ext}$ are the external force and torque acting on the vehicle. For aerial manipulators, the dynamic effect caused by the motion of the manipulator on the quadrotor is treated as external disturbance.

The total thrust and moment can be written according to the angular velocities of the four rotors ($\Omega_{i}$) as follows:

\begin{equation}
\left\{\begin{array}{l}
f_{u}=k_{b}\left(\Omega_{1}^{2}+\Omega_{2}^{2}+\Omega_{3}^{2}+\Omega_{4}^{2}\right) \\
M_{u,1}=k_{b} d\left(\Omega_{4}^{2}-\Omega_{2}^{2}\right)\\
M_{u,2}=k_{b} d\left(\Omega_{3}^{2}-\Omega_{1}^{2}\right)\\
M_{u,3}=k_{d}\left(\Omega_{2}^{2}+\Omega_{4}^{2}-\Omega_{1}^{2}-\Omega_{3}^{2}\right)
\end{array}\right.
\end{equation}
where $d$ is the quadrotor's arm length, $k_{b}$ and $k_{d}$ stand for the thrust factor and the drag factor, respectively.  

\section{Controller Design} 

\begin{figure}
	\centering
		\includegraphics[width=0.43\textwidth]{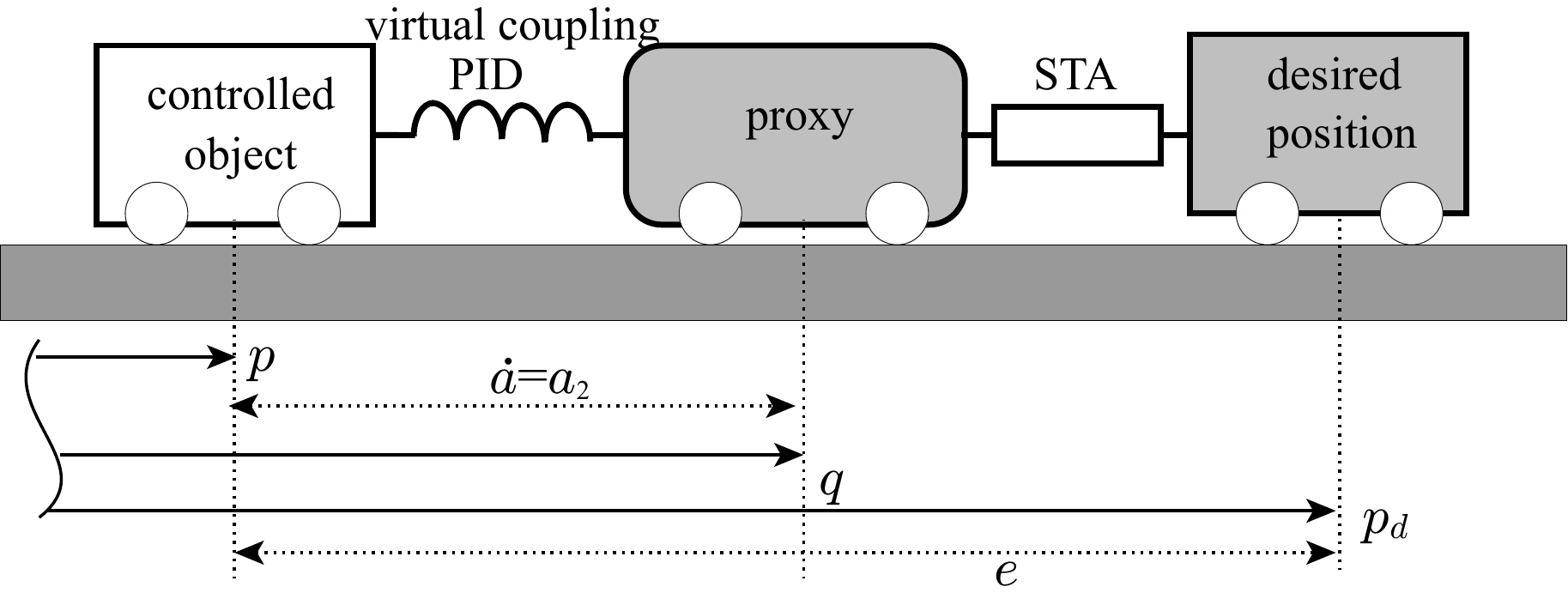}
	\caption{Illustration of the proposed proxy-based terminal sliding mode control}
	\label{Fig_Proxy_system}
\end{figure}
This section aims to design control input $f_{u}$ and $\boldsymbol{M_u}$ for position tracking control. We built a quadrotor controller in a cascaded structure by employing a proxy-based super-twisting algorithm as a kernel. As shown in Fig. 2, the entire controller diagram is depicted. For a given translational desired position $x_{d}$, a total thrust $f$ and a desired linear acceleration $\boldsymbol{a}_d$ are calculated by the outer loop's PSTA position controller. The desired attitude $R_d$ are determined by $\boldsymbol{a}_d$ and yaw command $\psi_d$, and the inner loop's PSTA attitude controller will compute a total torque $\boldsymbol{M_u}= [M_{u,1}, M_{u,2}, M_{u,3}]^{T}$ for quadrotor to approach the desired attitude. 

\begin{figure}[ht]
    \centering
    \includegraphics[width=0.43\textwidth]{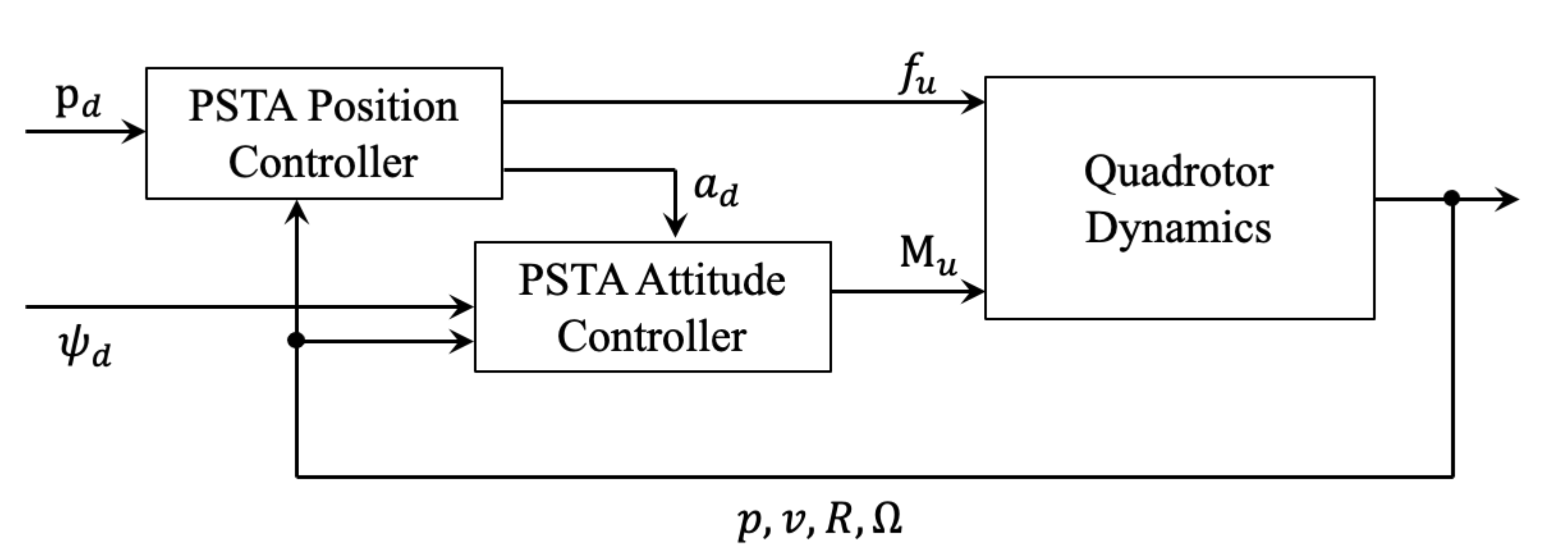}
    \caption{Controller Diagram}
    \label{Controller Diagram}
\end{figure}

\subsection{Related work: proxy-based sliding mode control (PSMC)}
\label{related_work}
The classical proxy-based sliding mode control was proposed by Kikuuwe et al. in \cite{Kikuuwe_2010_PSMC}, which is combination of the first-order SMC with the PID control:
\begin{subequations}\label{PSMC_Continuous_Time}
\begin{flalign}
& \sigma_1=p_{d}-p+H\left(\dot{p}_{d}-\dot{p}\right) \\
& 0=L a_1 +K \dot{a}_1+B \ddot{a}_1-F \operatorname{sgn}(\sigma_1-\dot{a}_1-H \ddot{a}_1) \\
& u=L a_1 +K \dot{a}_1+B \ddot{a}_1
\end{flalign}
\end{subequations}
 where $\sigma \in \bbR$ is the sliding variable, $p_d$ and $p$ are the desired position and the actual position of the controlled plant, as shown in Fig.~\ref{Fig_Proxy_system}, $H>0$ is a constant parameter of convergence rate, and $a$ is an intermediate variable defined as $\dot a_1=q-p$. The constant parameters, $L$, $K$, and $B$, are the integration, proportional, and derivative gains of the PID control, respectively. When $L=0$, Equation \eqref{PSMC_Continuous_Time} changes into a simplified version of PSMC with a combination of PD and SMC control \cite{Kikuuwe_2010_PSMC}.

\par

The PSMC has been widely employed for various applications\cite{KIKUUWE_European}, due to the characteristics of smooth and non-overshooting converging motion when actuators are saturated. One of the limitations of PSMC is the asymptotical accuracy of first-order SMC. This can be understood by rewriting Equation \eqref{PSMC_Continuous_Time} as a linear system of first-order with $L=0$ as follows:
\begin{flalign} \label{PSMC_newform}
& \dot a_2=-\frac{K}{B}a_2+\frac{F}{B}\sgn(\sigma_1-a_2-H\dot a_2)
\end{flalign}
where $a_2=\dot a_1$ and $\sigma$ can be viewed as a given reference. Therefore, one can consider Equation \eqref{PSMC_newform} as one linear first-order system controlled by the first-order SMC, and the term $-a_2{K}/{B}$ is the model part of the first-order system. Furthermore, the anti-windup property of the PSMC can be contributed to the feedback of actuation, i.e., $\sgn(\sigma-a_2-H\dot a_2)=\sgn(K\sigma-u)$ by setting $H=B/K$ and $L=0$.

\subsection{Proposed proxy-based super-twisting algorithm (PSTA)}
\label{pop_work}

\subsubsection{Continuous-time expression of PSTA}
To improve the control accuracy and inherit the properties of PSMC, let us employ the conditional super-twisting algorithm (STA) \cite{Coditional_STA_2021} instead of the first-order SMC:
\begin{subequations} \label{equ:u01_law}
\begin{flalign}
&\dot a_2=-\frac{K}{B}a_2+\frac{F_1}{B} \lfloor \sigma_1-a_2-H\dot a_2\rceil^{1/2} +v,  \label{equ:u01_lawa}  \\
& \dot v=\frac{F_2}{B}\lfloor  u^\ast-Bv \rceil^0,  u=Ka_2+B\dot{a}_2+F_2v, \label{equ:u01_lawb}
\end{flalign}
\end{subequations}
where $u^\ast=\mathrm{Proj}_\mathcal{A}(u)$, $\mathcal{A}=[-F,F]$ is the saturation level,  $F_1,F_2$ are two positive parameters, and $u^\ast$ is the input with $u^\ast=u$ when the saturation is inactive. The interpretation of the variable $a_2$ and parameters $K,B$ can be the same as in Equation \eqref{PSMC_Continuous_Time}.  The integral parameter $L=0$ is set to simplify the design and prevent numerical overﬂow of $a_1$ in Equation \eqref{PSMC_Continuous_Time}.
The notation $\lfloor \cdot \rceil^\xi$ is defined as $\forall x \in \bbR$, $\lfloor x \rceil^\xi=|x|^\xi \mathrm{sign}(x)$; $\forall \xi>0$, $\lfloor 0 \rceil^\xi=0$; $\lfloor x \rceil^0=\mathrm{sign}(x)$. The saturation function is defined as $ \forall x \in \bbR, \mathrm{Proj}_\mathcal{A}(x)=F\mathrm{sign}(x)$ if $|x|\geq F$ and  $\mathrm{Proj}_\mathcal{A}(x)=x$ if $|x|< F$. In \cite{DING_Automatica_2016}, with proper selections of $F_1$ and $F_2$ with the knowledge of the model parameters $K,B$, the closed-loop system Equation \eqref{equ:u01_law} is finite-time stable.

\par

The proxy-based super-twisting algorithm (PSTA) Equation \eqref{equ:u01_law} can be viewed as the linear system Equation \eqref{PSMC_newform}  with the first-order SMC replaced by the conditional STA. Therefore, the stability analysis of Equation \eqref{equ:u01_law} can be obtained directly by following the results in \cite{Coditional_STA_2021}.

\par

\par

It should be noted that if $|u^\ast|\leq F$, that is, the saturation is inactive,  from Equation \eqref{PSMC_Continuous_Time}, one has $u^\ast=u$ and
\begin{subequations} \label{equ:u02_law}
\begin{flalign}
&\dot a_2=-\frac{K}{B}a_2+\frac{F_1}{B} \lfloor \sigma_1-a_2-H\dot a_2\rceil^{1/2} +v,  \label{equ:u02_lawa}  \\
& \dot v=\frac{F_2}{B}\lfloor   \sigma_1-a_2-H\dot a_2 \rceil^0,  u=Ka_2+B\dot{a}_2+F_2 v. \label{equ:u02_lawb}
\end{flalign}
\end{subequations}

\subsubsection{Discretization scheme of PSTA}
\label{sec:one_stage}
In this realization scheme, we first rewrite the system Equation \eqref{equ:u01_law} as follows:
\begin{subequations} \label{equ:nominal_01}
\begin{flalign}
&\dot a_2=-ca_2+\kappa_1 \lfloor \sigma_1-\sigma_2 \rceil^{\frac{1}{2}} +v,  \label{equ:nominal_01a}   \\
& \dot v\in \kappa_2\lfloor  u^\ast-Bv \rceil^0,  u=Ka_2+B\dot{a}_2+F_2v, \label{equ:nominal_01b}
\end{flalign}
\end{subequations}
where $c:=K/B$, $\kappa_1:=F_1/B$, $\kappa_2:=F_2/B$, and $\sigma_2:=a_2+H\dot{a}_2$.
Let us discretize Equation \eqref{equ:nominal_01} with the implicit Euler method:
 \begin{subequations} \label{equ:nominal_02}
\begin{flalign}
&\frac{a_{2,k}-a_{2,k-1}}{h} \!=\! -ca_{2,k}+\kappa_1 \lfloor \sigma_{1,k}\!-\sigma_{2,k} \rceil^{\frac{1}{2}}\!+v_{k}  \label{equ:nominal_02a} \\
&\frac{v_{k}-v_{k-1}}{h} \in \kappa_2\lfloor  u^\ast_{k}-Bv_{k} \rceil^0,  u^\ast_{k}=\mathrm{Proj}_\mathcal{A}(u_k)  \label{equ:nominal_02b}
\end{flalign}
\end{subequations}
where $u_k=Ka_{2,k}+B\dot{a}_{2,k}+F_2v_k$ and $h$ is the time stepping size. Rearranging Equation \eqref{equ:nominal_02a} leads to the following expression:
\begin{flalign}\label{equ:discrete_001}
&(1+hc)a_{2,k} \!=\! \kappa_1 \lfloor h\sigma_{1,k}+Ha_{2,k-1}-(H+h)a_{2,k} \rceil^{\frac{1}{2}}\nonumber &\\
&+a_{2,k-1}+hv_{k}.  
\end{flalign}
If $|u^\ast|\leq F$ and $u^\ast=u$, Equation \eqref{equ:nominal_02b} can be equivalently written as:
\begin{flalign}\label{equ:discrete_002}
&v_k \!\in \! v_{k-1}+\kappa_2 \lfloor  h\sigma_{1,k}+Ha_{2,k-1}-(H+h)a_{2,k}  \rceil^0.
\end{flalign}
Substituting $v_k$ in Equation \eqref{equ:discrete_001} with Equation \eqref{equ:discrete_002} leads to
\begin{flalign}\label{equ:discrete_003}
&y_k\!\in \! \rho_{k-1}-\lambda_1\kappa_1 \lfloor y_k \rceil^{\frac{1}{2}}\!-\lambda_1 h \kappa_2 \lfloor  y_k  \rceil^0
\end{flalign}
where $y_k:=(H+h)a_{2,k} -(h\sigma_{1,k}+Ha_{2,k-1})$, $\rho_{k-1}:=\lambda_1( a_{2,k-1}+hv_{k-1})$ and $\lambda_1:=(H+h)/(1+hc)$. From Equation \eqref{equ:discrete_003}, one has the conclusion that
the relation $|y_k|\leq |\rho_k|$ between the sequences $\{y_k\}$ and $\{\rho_{k-1}\}$ holds for all $k\geq 1$. Therefore, from Equation \eqref{equ:discrete_003}, one has
\begin{flalign}\label{equ:x2star}
&y_k\in \rho_{k-1}\!-\! \kappa_3\sgn(y_k)\Leftrightarrow   \! y_{k}\!=\!\rho_{k-1}\!-\!\mathrm{Proj}_\mathcal{B}(\rho_{k-1})  &
\end{flalign}
where $\mathcal{B}=[-\kappa_3,\kappa_3]$, $\kappa_3:=\lambda_1(\kappa_1|\rho_{k-1}|^{\frac{1}{2}}+h\kappa_2)$, and the properties $x\in |y_k|^{1/2}\sgn(y_{k})\Rightarrow x \in |\rho_{k-1}|^{1/2} \sgn(y_{k})$ for $x\in \bbR$ and $|y_k|\leq |\rho_{k-1}|$ and $\forall x,y\in \bbR, \alpha>0, x=\kappa_3 \sgn(y-x)\Leftrightarrow x=\mathrm{Proj}_\mathcal{B}(y)$ in \cite{Xiong_2019_TCASII} have been used.
Finally, the proposed integration scheme of Equation \eqref{equ:nominal_01} is as follows for the case $u^\ast=u$:
 \begin{subequations} \label{equ:sat_inact}
\begin{flalign}
&  a_{2,k}=\lambda_2z_{k}-\lambda_2 \mathrm{Proj}_\mathcal{B}(\rho_{k-1})  \\
& v_k =  v_{k-1}-\dfrac{\kappa_2}{\kappa_3} \mathrm{Proj}_\mathcal{B}(\rho_{k-1})
\end{flalign}
\end{subequations}
where $z_{k}:=h \sigma_{1,k}+Ha_{2,k-1}+\rho_{k-1}$ and $\lambda_2:=1/(H+h)$. Then, one has $u_k=Ka_{2,k}+B(a_{2,k}-a_{2,k-1})/h$ as the control input.

\par

Now let us consider the case $u^\ast\neq u$, that is, the saturation is active. From Equation \eqref{equ:nominal_02b}, one has
\begin{flalign}\label{sat_active}
  &v_{k} \in v_{k-1}+ h\kappa_2\lfloor  u^\ast_{k}/B-v_{k} \rceil^0,  u^\ast_{k}=\mathrm{Proj}_\mathcal{A}(u_k),
\end{flalign}
which can be further equivalently transferred into the following equation:
\begin{flalign}\label{sat_active02}
  &v_{k} = v_{k-1}- \mathrm{Proj}_\mathcal{C}\left(v_{k-1}-\frac{ u^\ast}{B}\right),  u^\ast_{k}=\mathrm{Proj}_\mathcal{A}(u_k),
\end{flalign}
with $\mathcal{C}:=[-h\kappa_2, h\kappa_2]$ and $u_k$ is calculated from Equation \eqref{equ:sat_inact}.

\par

Finally, the proposed discrete-time implementation scheme for Equation \eqref{equ:nominal_01} can be summarized as follows:
 \begin{subequations} \label{nominal_imp_lemma}
 \begin{flalign}
 &\lambda_1=(H+h)/(1+hc), \lambda_2=1/(H+h), & \\
 &\rho_{k-1}:=\lambda_1( a_{2,k-1}+hv_{k-1}), u^\ast_{k}=\mathrm{Proj}_\mathcal{A}(u_k)  & \\
 & z_{k}:=h \sigma_{1,k}+Ha_{2,k-1}+\rho_{k-1} & \\
 &a_{2,k}=\lambda_2z_{k}-\lambda_2 \mathrm{Proj}_\mathcal{B}(\rho_{k-1})  & \\
 &  u_k=Ka_{2,k}+B(a_{2,k}-a_{2,k-1})/h+v_k, & \\
 & v_k\!=\! \begin{cases}  v_{k-1}-\dfrac{\kappa_2}{\kappa_3} \mathrm{Proj}_\mathcal{B}(\rho_{k-1})  \hspace{-0.5em} &\mbox{if }|u_k|\!\leq F \\
 v_{k-1}- \mathrm{Proj}_\mathcal{C}(v_{k-1}-u^\ast/B) \hspace{-0.5em} & \mbox{else } \end{cases}
 \end{flalign}
\end{subequations}
with $c=K/B$.

\subsection{Implementation on Quadrotor}

For convenience, we wrapped Equation \eqref{nominal_imp_lemma} as $
{PSTA}_{B, K, H}(\sigma_{1,k})$, a function of the sliding mode variable $\sigma_{1,k}$, which outputs the control command $u_k$. The function also includes the update step of intermediate terms $a_{2,k}$ and $v_k$. Given the desired position and yaw angle, the desired attitude and thrust are calculated through the PSTA position controller in the outer loop. Desired acceleration at each axis in the world frame $\boldsymbol{a}_{desired}$ are calculated by PSTA :

\begin{equation}
\boldsymbol{a}_{desired}=\left[\begin{array}{c}a_{{x}} \\ a_{{y}} \\ a_{{z}}\end{array}\right]=\left[\begin{array}{c}PSTA_{B_x, K_x, H_x}(\sigma_{x}) \\ PSTA_{B_y, K_y, H_y}(\sigma_{y}) \\ PSTA_{B_z, K_z, H_z}(\sigma_{z}) \end{array}\right],
\end{equation}
where $\sigma_{x}$, $\sigma_{y}$, $\sigma_{z}$ are the sliding variables:

\begin{equation}
\boldsymbol{\sigma}_{tran}=\left[\begin{array}{c}\sigma_{{x}} \\ \sigma_{{y}} \\ \sigma_{{z}}\end{array}\right]=\left[\begin{array}{c}x_{d}-x+H\left(\dot{x}_{d}-\dot{x}\right)\\ y_{d}-y+H\left(\dot{y}_{d}-\dot{y}\right) \\ z_{d}-z+H\left(\dot{z}_{d}-\dot{z}\right) \end{array}\right],
\end{equation}
while the thrust command $f_u$ can be derived as:
\begin{equation}
f_u=m\boldsymbol{a}_{desired} \cdot R e_3,
\end{equation}
where the unit vector $e_3=\left[\begin{array}{lll}0 & 0 & 1\end{array}\right]^{T}$.

Following a similar approach as in \cite{lee2010geometric}, we construct the desired rotation matrix $\boldsymbol{R}_d$ as described below. Firstly, the desired third body frame axis is given by:

\begin{equation}
\mathbf{b}_{3 d}=\frac{\boldsymbol{a}_{desired}}{\left\|\boldsymbol{a}_{desired}\right\|}, \boldsymbol{a}_{desired} \neq \mathbf{0}.
\end{equation}
The first body frame is defined initially in the xy plane, then it is projected on the plane perpendicular to $\mathbf{b}_{3 d}$. Hence  $\mathbf{b}_{1 d}$ is given by:
\begin{equation}
\mathbf{b}_{1 d}=\left[\begin{array}{lll}\cos \psi_{d} & \sin \psi_{d} & 0\end{array}\right]^{T},
\end{equation}
where $\psi_{d}$ represents the desired yaw angle. The second body frame is perpendicular to the plane constructed by $\mathbf{b}_{3 d}$ and $\mathbf{b}_{1 d}$:
\begin{equation}
\mathbf{b}_{2 d}=\frac{\mathbf{b}_{3 d} \times \mathbf{b}_{1 d}}{\left\|\mathbf{b}_{3 d} \times \mathbf{b}_{1 d}\right\|}.
\end{equation}
Finally, the desired rotation matrix $\boldsymbol{R}_d$ is given by:
\begin{equation}
\mathbf{R}_{d}=\left[\begin{array}{lll}\mathbf{b}_{2 d} \times \mathbf{b}_{3 d} & \mathbf{b}_{2 d} & \mathbf{b}_{3 d}\end{array}\right].
\end{equation}
The attitude tracking error $\boldsymbol{e}_{{R}} \in \mathbb{R}^{3}$ is defined as:
\begin{equation}
\mathbf{e}_{R}=\frac{1}{2}\left(\mathbf{R}_{d}^{T} \mathbf{R}-\mathbf{R}^{T} \mathbf{R}_{d}\right)^{\vee},
\end{equation}
while the angular velocity tracking error is described as follow: 
\begin{equation}
\mathbf{e}_{\omega}=\boldsymbol{\omega}-\mathbf{R}^{T} \mathbf{R}_{d} \boldsymbol{\omega}_{d},
\end{equation}
where $[\cdot]_{\vee}$ represents the inverse (vee) operator from $so(3)$ to $\mathbb{R}^{3}$. 
With the definition of angular and angular velocity error, the sliding variable of rotation is:
\begin{equation}
 \mathbf{\sigma}_{rot}=\left[\begin{array}{c}\sigma_{{\theta}} \\ \sigma_{{\phi}} \\ \sigma_{{\psi}}\end{array}\right]=\left[\begin{array}{c}\mathbf{e_R(1)}+H \mathbf{e_\omega(1)}\\ \mathbf{e_R(2)}+H \mathbf{e_\omega(2)} \\ \mathbf{e_R(3)}+H \mathbf{e_\omega(3)} \end{array}\right].
\end{equation}

The attitude controller in the outer loop utilized PSTA again to calculate the control inputs: 
\begin{equation}
\boldsymbol{M}_{u}=\left[\begin{array}{c}M_{{u,1}} \\ M_{{u,2}} \\ M_{{u,3}}\end{array}\right]=\left[\begin{array}{c}J_x \cdot PSTA_{B_\theta, K_\theta, H_\theta}(\sigma_{\theta})\\ J_y \cdot PSTA_{B_\phi, K_\phi, H_\phi}(\sigma_{\phi})\\ J_z \cdot PSTA_{B_\psi, K_\psi, H_\psi}(\sigma_{\psi}) \end{array}\right].
\end{equation}

\section{Numerical Simulation Results}

\begin{figure}[ht]
    \centering
    \includegraphics[width=9.0cm]{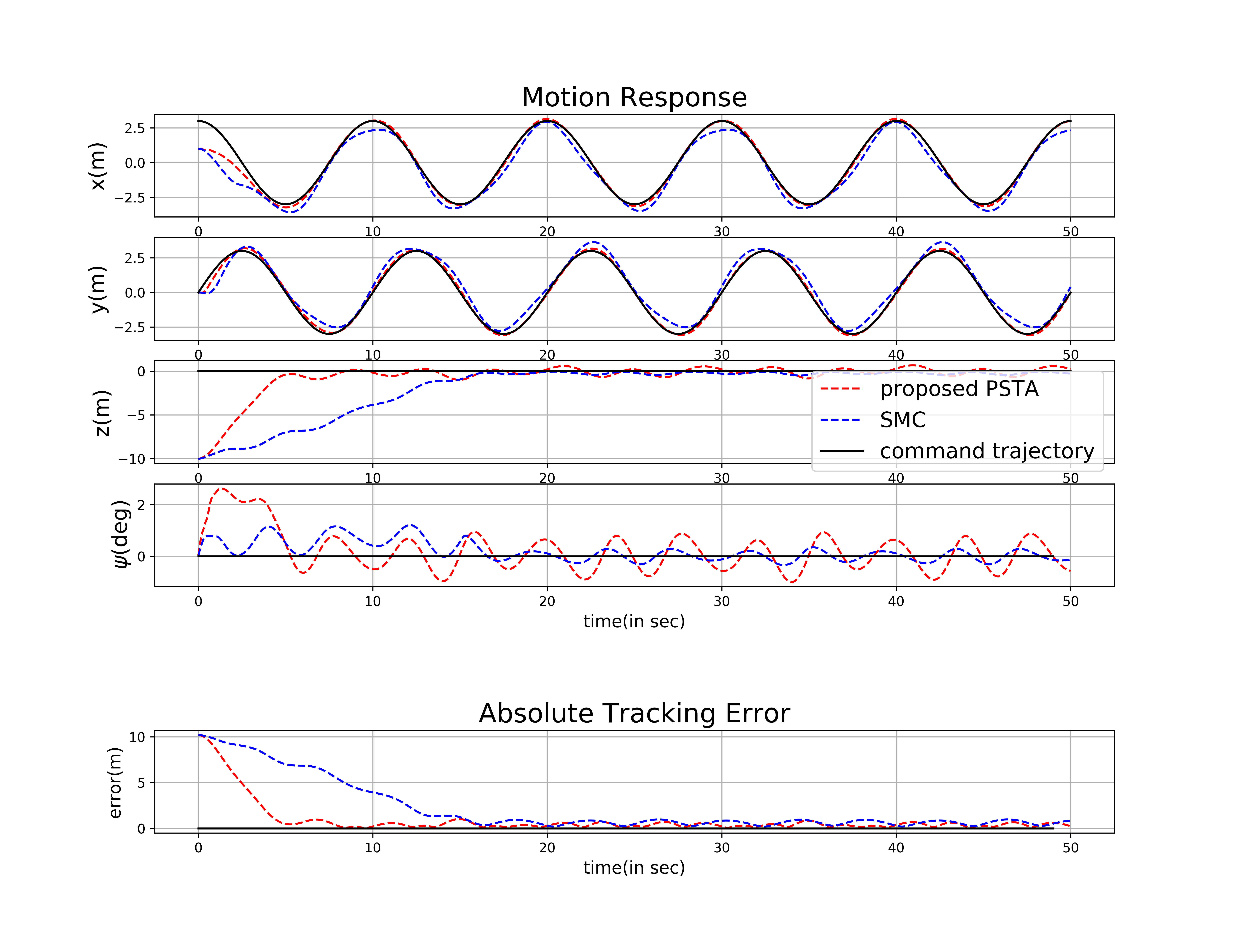}
    \caption{Position and yaw angle responses, and absolute tracking error.}
    \label{psta_test}
\end{figure}

\begin{figure}[ht]
    \centering
    \includegraphics[width=9.0cm, height=6.0cm]{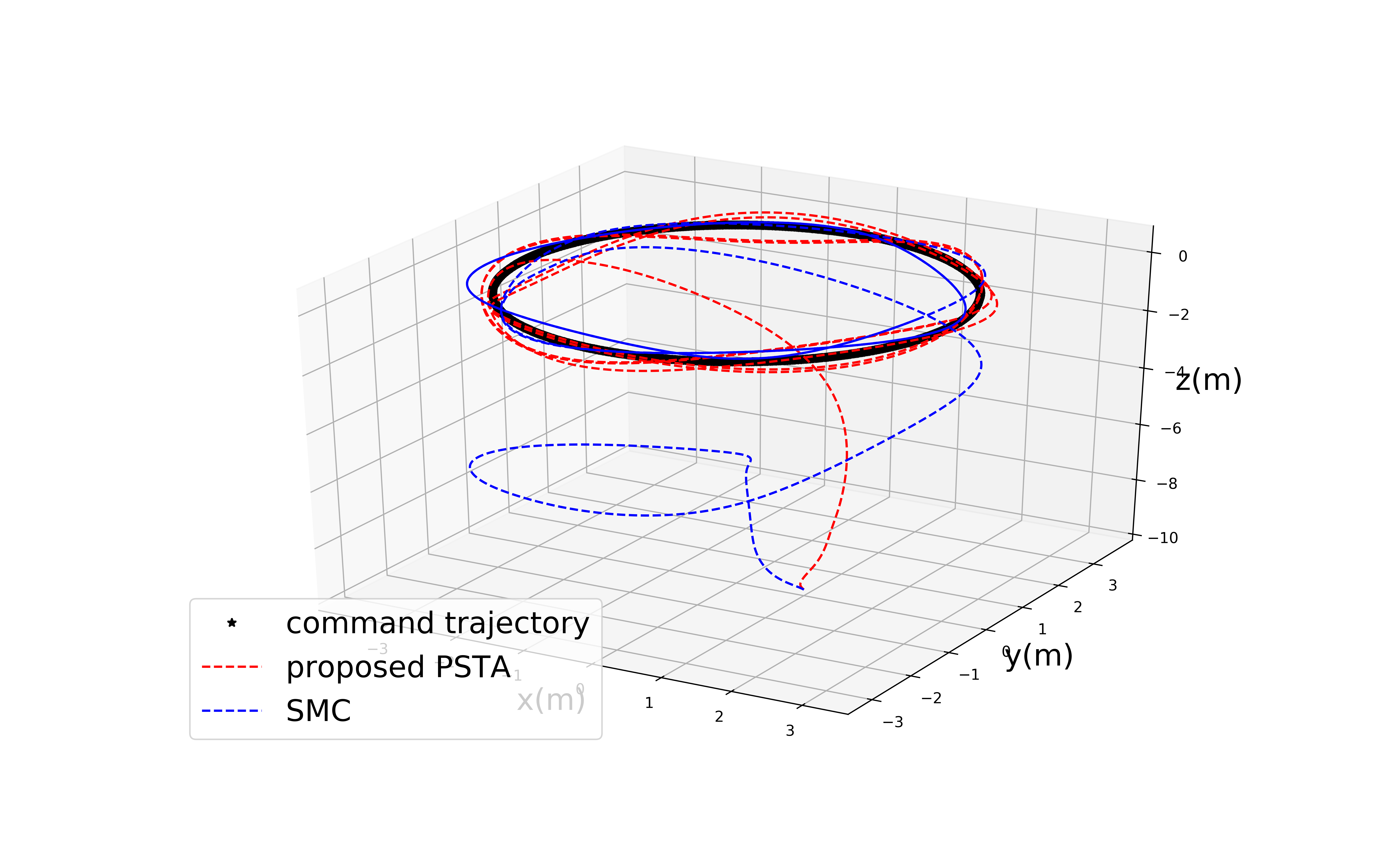}
    \caption{3D Trajectory of the Quadrotor.}
    \label{3Dtraj}
\end{figure}

\begin{figure}[ht]
    \centering
    \includegraphics[width=9.0cm]{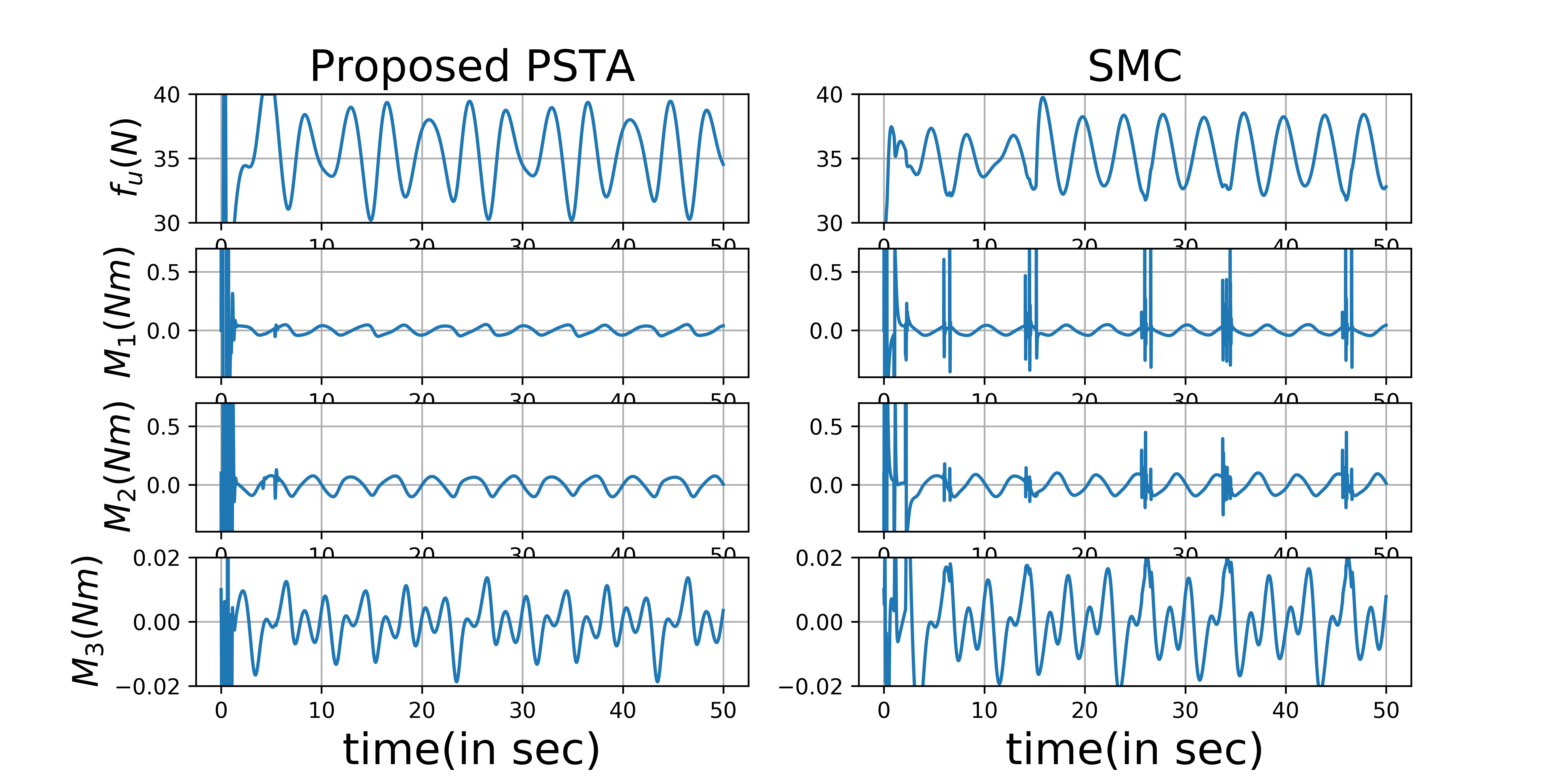}
    \caption{Control inputs $f_u$ and $\boldsymbol{M_u}$ based on PSTA and conventional SMC\cite{l2018introduction}..}
    \label{input}
\end{figure}

This section presents the numerical simulation results for the proposed PSTA quadrotor controller. 
The simulation is conducted based on the open source quadrotor numerical platform nonlinearquad\cite{nonlinearquad}. To show the effectiveness of proposed PSTA controller, simulation results are compared with a conventional sliding mode control approach\cite{l2018introduction}, which is already implemented and well-tuned by the contributor of the open source project.

Default simulation parameters were set as: $m=3.81 \mathrm{kg}, J_{x}= 0.1 \mathrm{~kg} . \mathrm{m}^{2}, J_{y}=0.1\mathrm{~kg} \cdot \mathrm{m}^{2}, J_{z}=0.1 \mathrm{~kg} \cdot \mathrm{m}^{2}$.  
The task of the simulation is to make the quadrotor track a circular trajectory of a radius 1m. The initial states of quadrotor are given as $\mathbf{x}_0=[x_0, \dot{x}_0, y_0, \dot{y}_0, z_0, \dot{z}_0, \theta_0, \dot{\theta}_0, \phi_0, \dot{\phi}_0,  \psi_0, \dot{\psi}_0]^{T}=[0,0,-10,0,0,0,0,0,0,0,0,0,0]^T$. The desired position and yaw angle are set as $x_{d}=\cos (2\pi\cdot0.1 t), y_{d}=\sin (2\pi\cdot0.1 t), z_{d}=0 \text { and } \psi_{d}=0$. For disturbances simulation, we set the external forces $\boldsymbol{f}_{ext}$ and torques $\boldsymbol{M}_{ext}$  as below:

\begin{equation}
\boldsymbol{f}_{ext}=\left[\begin{array}{c}f_{ext, x} \\ f_{ext, y} \\ f_{ext, z}\end{array}\right]=\left[\begin{array}{c}10\sin (2\pi\cdot0.25 t)\\ 10\cos (2\pi\cdot0.25 t) \\ 0.1\cos (2\pi\cdot0.25 t)\end{array}\right] (N)
\end{equation}

\begin{equation}
\boldsymbol{M}_{ext}=\left[\begin{array}{c}M_{ext, 1} \\ M_{ext, 2} \\ M_{ext, 3}\end{array}\right]=\left[\begin{array}{c}0.1\sin (2\pi\cdot0.25 t)\\ 0.1\cos (2\pi\cdot0.25 t) \\ 0.1\cos (2\pi\cdot0.25 t)\end{array}\right] (N\cdot m)
\end{equation}

Fig.~\ref{psta_test} shows the simulated quadrotor tracking results using the control approaches of conventional sliding mode control and our proposed PSTA, respectively. Fig.~\ref{3Dtraj} presents the results in 3D plot. Fig.~\ref{input} shows the control efforts $f_u$ and $\boldsymbol{M}_{u}$. The proposed PSTA controller outperformed conventional SMC. And even in such a challenging scenario with strong disturbances, it provided outstanding tracking performance and suppressed the chattering effectively. 

\section{Gazebo Simulation Results}

\begin{figure}[t]
    \begin{minipage}[t]{3.5cm}
    \centering
    \includegraphics[width=\textwidth]{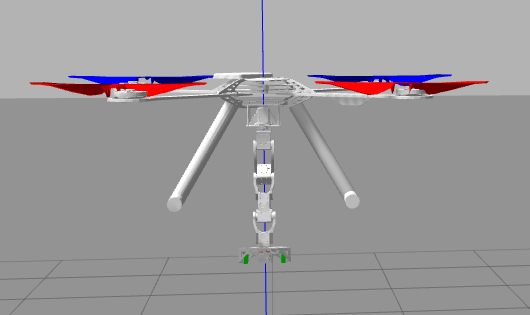}
    \end{minipage}
    \hspace{1.48mm}
    \begin{minipage}[t]{4.0cm}
    \centering
    \includegraphics[width=\textwidth]{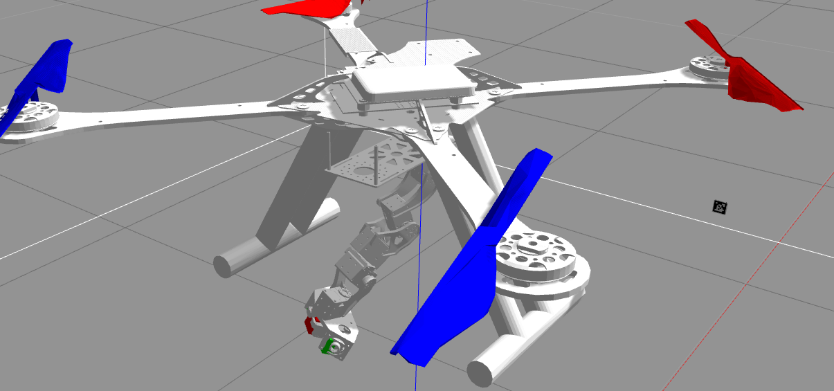}
    \end{minipage}      
    \caption{Gazebo simulation platform: hummingbird quadrotor with a 6-DOF manipulator.}
    \label{fig_fullsystem}
\end{figure}

\begin{figure}[ht]
    \centering
    \includegraphics[width=7.0cm,height=5.0cm]{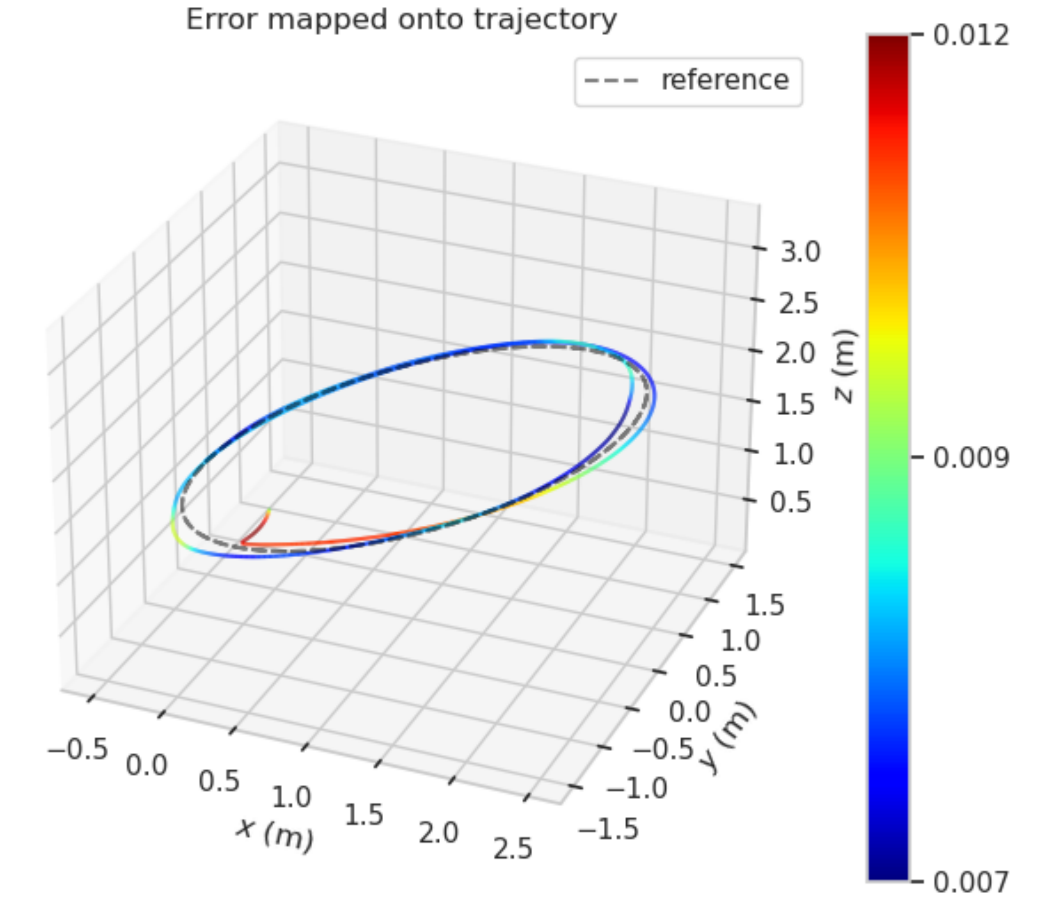}
    \caption{3D Trajectory of the Quadrotor.}
    \label{gazebo_3d}
\end{figure}
\begin{figure}[ht]
    \centering
    \includegraphics[width=7.0cm,height=5.0cm]{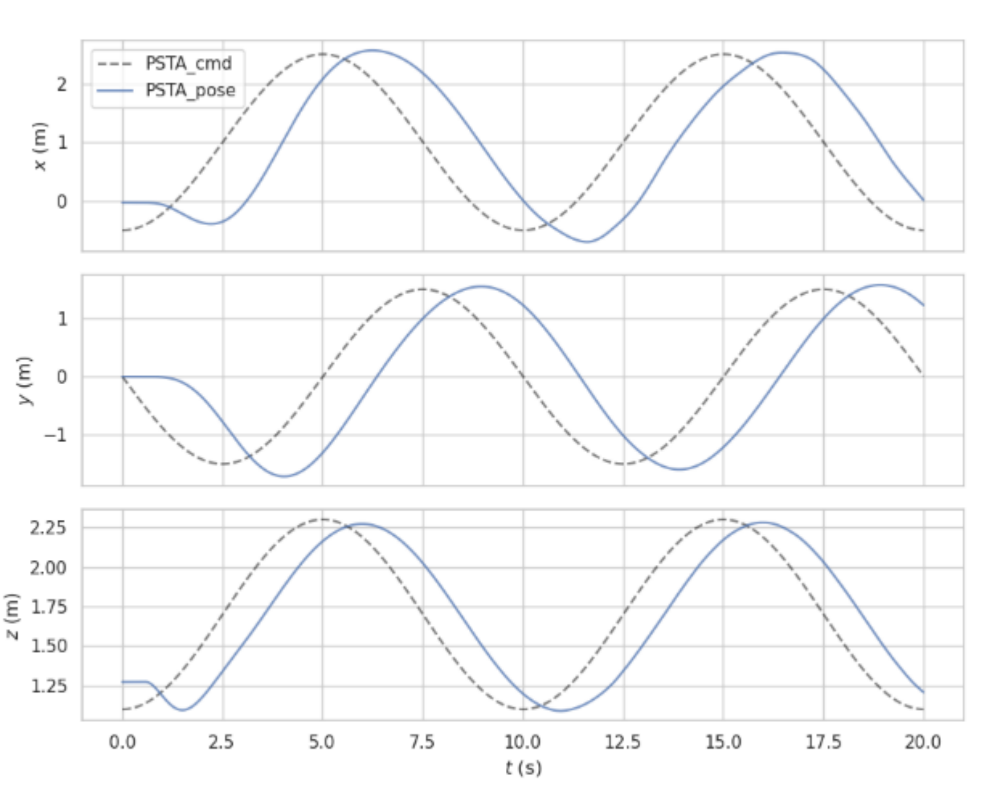}
    \caption{Position tracking result.}
    \label{gazebo_xyz}
\end{figure}
\begin{figure}[ht]
    \centering
    \includegraphics[width=7.0cm,height=5.0cm]{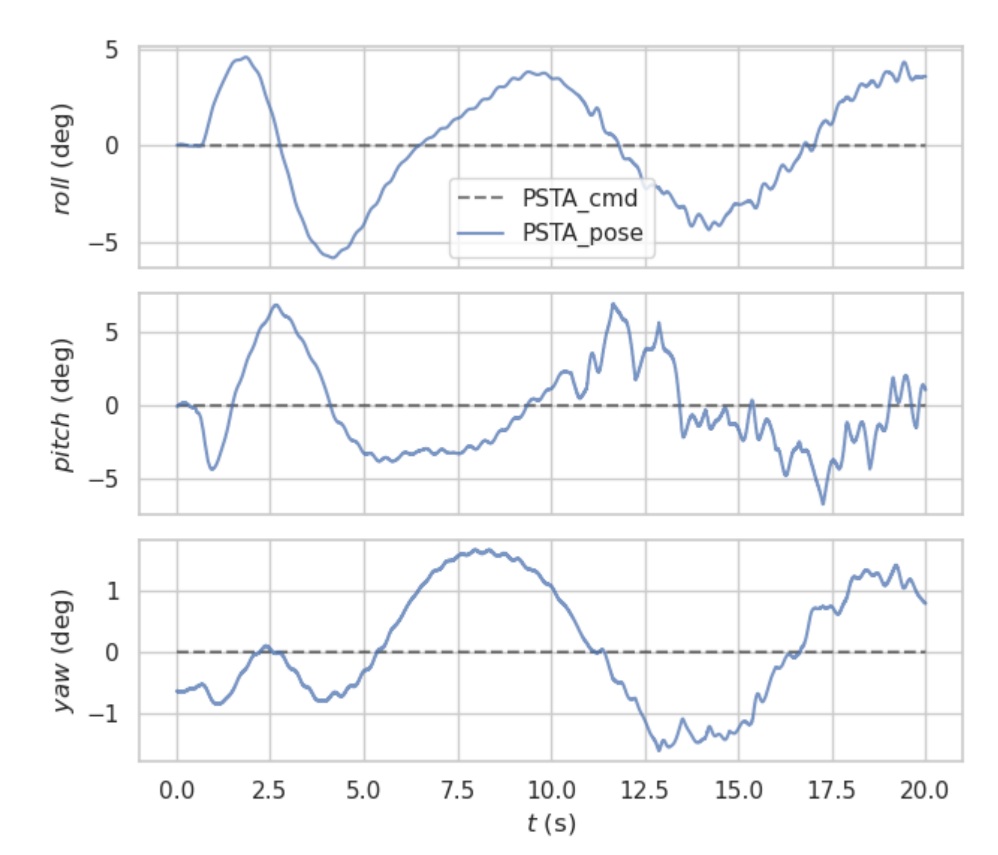}
    \caption{Attitude tracking result.}
    \label{gazebo_rpy}
\end{figure}
\begin{figure}[ht]
    \centering
    \includegraphics[width=8.0cm,height=5.0cm]{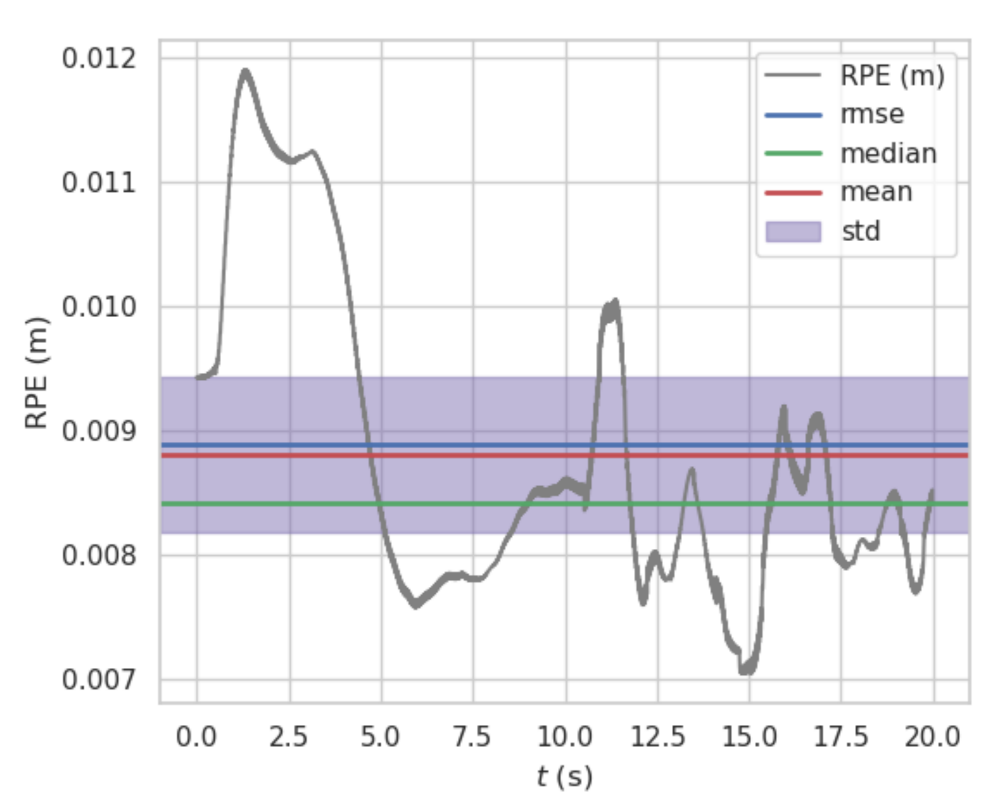}
    \caption{Relative pose error in meters.} 
    \label{gazebo_rpe}
\end{figure}

Besides numerical experiments that test the quadrotor controller under disturbances, we also evaluated the proposed controller in Gazebo simulation environment.  As shown in Fig~\ref{fig_fullsystem}, the aerial manipulation system is built based on the quadrotor model named hummingbird in RotorS\cite{Furrer2016} by mounting a 6-DOF manipulator. Quadrotor parameters were given as: $m=0.7 \mathrm{kg}, J_{x}= 0.007 \mathrm{kg} . \mathrm{m}^{2}, J_{y}=0.007\mathrm{~kg} \cdot \mathrm{m}^{2}, J_{z}=0.012 \mathrm{~kg} \cdot \mathrm{m}^{2}, d_{arm}=0.17\mathrm{m}$.  The The total mass and length of the mounted manipulator is about $0.2 \mathrm{kg}$ and $~0.2 \mathrm{m}$, respectively. 

In the test case, the reference elliptical trajectory is set as:
\begin{equation}
\begin{aligned} x_{d} &=-1.5 \cos (2\pi \cdot 0.2t) \\ y_{d} &=1.0-1.5 \sin (2\pi \cdot 0.2t) \\z_{d}&=1.6-0.6 \sin (2\pi \cdot 0.2t) \\ \psi_{d} &=0 \end{aligned}
\end{equation}

The task of this case is to make the quadrotor track the elliptical trajectory for two periods. The robot manipulator remains stable in the first cycle, and in the second cycle, a sinusoidal control signal is sent to its second joint and makes it move. The motion of the manipulator generates external forces and torques to the quadrotor. Results are presented in Figures ~\ref{gazebo_3d}, \ref{gazebo_xyz}, \ref{gazebo_rpy} and \ref{gazebo_rpe}. Even under consistent disturbances, the results show that the proposed controller can help the quadrotor achieve a satisfying tracking performance with a relative position error lower than 9mm.

We also conducted a comparison between the proposed controller and a disturbance rejection model predictive controller\cite{Hentzen2019} in Gazebo. To ensure fairness in the comparison, we subjected both controllers to testing using the open-source project attached with \cite{Hentzen2019}. All parameters of MPC were set to their default values. The PSTA controller that we proposed demonstrated superior performance to the MPC in the control group. For space reasons, in-depth results have been provided in the accompanying video.

\section{CONCLUSIONS}

We present a proxy-based super twisting algorithm that has been implemented on a quadrotor aerial manipulator. The motion of the manipulator produces disturbances in the quadrotor UAV. While the super twisting sliding mode algorithm can alleviate these disturbances, but introduces chattering effect. Our proposed proxy-based super twisting algorithm mitigates this issue by incorporating a proxy state. It has been further developed as a cascaded PSTA controller to track the quadrotor's position. Both numerical and Gazebo simulations have been conducted to affirm the effectiveness of the controller. In the future, parameter tuning and disturbances estimation will be explored to enhance the controller's tracking accuracy. Additionally, real-world experiments will be performed to establish the viability of the proposed control method.




\bibliographystyle{unsrt}

\begin{thebibliography}{10}

\bibitem{RuggieroSurvey2018}
Fabio Ruggiero, Vincenzo Lippiello, and Anibal Ollero.
\newblock {Aerial Manipulation: A Literature Review}.
\newblock {\em IEEE Robotics and Automation Letters}, 3(3):1957--1964, 2018.

\bibitem{ruggiero2014impedance}
Fabio Ruggiero, Jonathan Cacace, Hamid Sadeghian, and Vincenzo Lippiello.
\newblock Impedance control of vtol uavs with a momentum-based external
  generalized forces estimator.
\newblock In {\em 2014 IEEE international conference on robotics and automation
  (ICRA)}, pages 2093--2099. IEEE, 2014.

\bibitem{ruggiero2015passivity}
Fabio Ruggiero, Jonathan Cacace, Hamid Sadeghian, and Vincenzo Lippiello.
\newblock Passivity-based control of vtol uavs with a momentum-based estimator
  of external wrench and unmodeled dynamics.
\newblock {\em Robotics and Autonomous Systems}, 72:139--151, 2015.

\bibitem{jimenez2013control}
Antonio~E Jimenez-Cano, Jes{\'u}s Martin, Guillermo Heredia, An{\'\i}bal
  Ollero, and Raul Cano.
\newblock Control of an aerial robot with multi-link arm for assembly tasks.
\newblock In {\em 2013 IEEE International Conference on Robotics and
  Automation}, pages 4916--4921. IEEE, 2013.

\bibitem{kim2017robust}
Suseong Kim, Seungwon Choi, Hyeonggeun Kim, Jongho Shin, Hyungbo Shim, and
  H~Jin Kim.
\newblock Robust control of an equipment-added multirotor using disturbance
  observer.
\newblock {\em IEEE Transactions on Control Systems Technology},
  26(4):1524--1531, 2017.

\bibitem{xu2006sliding}
Rong Xu and Umit Ozguner.
\newblock Sliding mode control of a quadrotor helicopter.
\newblock In {\em Proceedings of the 45th IEEE Conference on Decision and
  Control}, pages 4957--4962. IEEE, 2006.

\bibitem{l2018introduction}
Andrea L'afflitto, Robert~B Anderson, and Keyvan Mohammadi.
\newblock An introduction to nonlinear robust control for unmanned quadrotor
  aircraft: how to design control algorithms for quadrotors using sliding mode
  control and adaptive control techniques [focus on education].
\newblock {\em IEEE Control Systems Magazine}, 38(3):102--121, 2018.

\bibitem{mofid2020adaptive}
Omid Mofid, Saleh Mobayen, and Wing-Kwong Wong.
\newblock Adaptive terminal sliding mode control for attitude and position
  tracking control of quadrotor uavs in the existence of external disturbance.
\newblock {\em IEEE Access}, 9:3428--3440, 2020.

\bibitem{Ding_2020_TCASI}
L.~{Liu}, W.~X. {Zheng}, and S.~{Ding}.
\newblock An adaptive sosm controller design by using a sliding-mode-based
  filter and its application to buck converter.
\newblock {\em IEEE Transactions on Circuits and Systems I: Regular Papers},
  67(7):2409--2418, 2020.

\bibitem{Utkin_2013_ASTA}
Vadim~I. Utkin and Alex~S. Poznyak.
\newblock Adaptive sliding mode control with application to super-twist
  algorithm: Equivalent control method.
\newblock {\em Automatica}, 49(1):39 -- 47, 2013.

\bibitem{Mobayen_2011_ACC}
S.~{Mobayen}, M.~J. {Yazdanpanah}, and V.~J. {Majd}.
\newblock A finite-time tracker for nonholonomic systems using recursive
  singularity-free {FTSM}.
\newblock In {\em Proceedings of the 2011 American Control Conference}, pages
  1720--1725, 2011.

\bibitem{Mobayen_2017_Scientia}
S.~Mobayen and F.~Tchier.
\newblock Nonsingular fast terminal sliding-mode stabilizer for a class of
  uncertain nonlinear systems based on disturbance observer.
\newblock {\em Scientia Iranica}, 24(3):1410--1418, 2017.

\bibitem{Brogliato_2019_STA}
B.~{Brogliato}, A.~{Polyakov}, and D.~{Efimov}.
\newblock The implicit discretization of the super-twisting sliding-mode
  control algorithm.
\newblock {\em IEEE Transactions on Automatic Control}, pages 1--1\,, 2019.

\bibitem{Xiong_2019_TCASII}
X.~{Xiong}, R.~{Kikuuwe}, S.~{Kamal}, and S.~{Jin}.
\newblock Implicit-euler implementation of super-twisting observer and twisting
  controller for second-order systems.
\newblock {\em IEEE Transactions on Circuits and Systems II: Express Briefs},
  67(11):2607--2611, 2020.

\bibitem{Brogliato_2020_differentiator}
Mohammad~Rasool Mojallizadeh, Bernard Brogliato, and Vincent Acary.
\newblock {Discrete-time differentiators: design and comparative analysis}.
\newblock hal-02960923, October 2020.

\bibitem{Kikuuwe_2010_PSMC}
R.~Kikuuwe, S.~Yasukouchi, H.~Fujimoto, and M.~Yamamoto.
\newblock {P}roxy-based sliding mode control: a safer extension of {PID}
  position control.
\newblock {\em {IEEE} Trans. on Robotics}, 26(4):670--683, 2010.

\bibitem{KIKUUWE_European}
Ryo Kikuuwe.
\newblock Sliding motion accuracy of proxy-based sliding mode control subjected
  to measurement noise and disturbance.
\newblock {\em European Journal of Control}, 58:114--122, 2021.

\bibitem{Coditional_STA_2021}
Richard Seeber and Markus Reichhartinger.
\newblock Conditioned super-twisting algorithm for systems with saturated
  control action.
\newblock {\em Automatica}, 116:108921, 2020.

\bibitem{DING_Automatica_2016}
Shihong Ding, Arie Levant, and Shihua Li.
\newblock Simple homogeneous sliding-mode controller.
\newblock {\em Automatica}, 67:22 -- 32, 2016.

\bibitem{lee2010geometric}
Taeyoung Lee, Melvin Leok, and N~Harris McClamroch.
\newblock Geometric tracking control of a quadrotor uav on se (3).
\newblock In {\em 49th IEEE conference on decision and control (CDC)}, pages
  5420--5425. IEEE, 2010.

\bibitem{nonlinearquad}
https://github.com/plusk01/nonlinearquad.

\bibitem{Furrer2016}
Fadri Furrer, Michael Burri, Markus Achtelik, and Roland Siegwart.
\newblock {\em Robot Operating System (ROS): The Complete Reference (Volume
  1)}, chapter RotorS---A Modular Gazebo MAV Simulator Framework, pages
  595--625.
\newblock Springer International Publishing, Cham, 2016.

\bibitem{Hentzen2019}
Daniel Hentzen, Thomas Stastny, Roland Siegwart, and Roland Brockers.
\newblock Disturbance estimation and rejection for high-precision multirotor
  position control.
\newblock In {\em 2019 IEEE/RSJ International Conference on Intelligent Robots
  and Systems (IROS)}, pages 2797--2804, 2019.

\end{thebibliography}
\end{document}